\documentclass{article} 
\usepackage{nips14submit_e,times}
\usepackage{hyperref}
\usepackage{url}

\usepackage{caption}
\usepackage{subcaption}
\usepackage{graphicx}
\usepackage{amsmath}
\graphicspath{ {images/} }

\title{Composable Unpaired Image to Image Translation}

\author{
Laura Graesser \\
New York University \\
\texttt{lhg256@nyu.edu} \\
\AND
Anant Gupta \\
New York University \\
\texttt{ag4508@nyu.edu} \\
}

\nipsfinalcopy 

\begin{document}

\maketitle

\begin{abstract}
There has been remarkable recent work in unpaired image-to-image translation. However, they're restricted to translation on single pairs of distributions, with some exceptions. In this study, we extend one of these works to a scalable multi-distribution translation mechanism. Our translation models not only converts from one distribution to another but can be stacked to create composite translation functions. We show that this composite property makes it possible to generate images with characteristics not seen in the training set. We also propose a decoupled training mechanism to train multiple distributions separately, which we show, generates better samples than isolated joint training. Further, we do a qualitative and quantitative analysis to assess the plausibility of the samples. The code is made available at \url{https://github.com/lgraesser/im2im2im}.
\end{abstract}

\section{Introduction}
Learning to translate between two image domains is a common problem in computer vision and graphics, and has many potentially useful applications including colorization \cite{pix2pix}, photo generation from sketches \cite{pix2pix}, inpainting \cite{inpainting}, future frame prediction \cite{framepredict}, superresolution \cite{superres}, style transfer \cite{style}, and dataset augmentation. It can be particularly useful when images from one of the two domains are scarce or expensive to obtain (for example by requiring human annotation or modification). \par
Until recently the problem has been posed as a supervised learning problem or a one-to-one mapping, with training datasets of paired images from each domain \cite{pix2pix}. However having access to paired images is a difficult and resource intensive challenge, and so it is helpful to learn to map between unpaired image distributions. Multiple approaches have been successfully applied in solving this task in the recent months \cite{cyclegan,cogan,unit,xgan,stargan}.

Most of the work done in this area deal with translation of images between a single pair of distributions. In this work, we generalize this translation mechanism to multiple pairs. In other words, given a set of distributions which share an underlying joint distribution, we come up with a set of translators that can convert samples from images belonging to any distribution to any other distribution, on which these translators were trained. The effectiveness of these translators is exhibited by considering them as a set of composite functions which can be applied on top of one another. 

Further, we explore if these models have the capability of disentanglement of shared and individual components between different distributions. For example, instead of learning to translate from a smiling person that is wearing glasses to a person that is not smiling and not wearing glasses, or a horse in a field on a summer's day to a zebra in a field on a winter's day, we learn to translate from wearing glasses to not wearing glasses, smiling to not smiling, horse to zebra, and summer to winter separately, then compose the results. \par


There are a number of potential advantages to this approach. It becomes possible to learn granular unpaired image to image translations, whilst only having access to either less granular or no labels. It facilitates training on larger datasets since only the marginal, more general labels are required. It gives finer grained control to users of the translation process since they can compose different translation functions to achieve their desired results. Finally, it makes it possible to generate entirely new combinations, by translating to combinations of the marginal distributions that never appeared in the training set.

We also experiment with different training mechanisms to efficiently train models on multiple distributions and show results on decoupled training performing better joint training. Overall, decoupled training followed by some finetuning by joint training produces the best results.

\section{Related Work}
\label{ref:related-work}

\textbf{Generative Adversarial Networks}: Image generation through GAN \cite{gan} and it's several variants such as DCGAN \cite{dcgan} and WGAN \cite{wgan} have been groundbreaking in terms of how realistic the generated samples were. The adversarial loss originally introduced in \cite{gan} has led to creation of a new kind of architecture in generative modelling and subsequently been applied in several areas such as \cite{pix2pix, inpainting, framepredict}. It consists of a generator and a discriminator, wherein the former learns to generate novel realistic samples in order to fool the latter, while the latter's objective is to distinguish between real samples and generated ones. The combined learning objective is to minimize the adversarial loss.

\textbf{Image-to-Image translation}: Supervised image-to-image translation \cite{pix2pix} has achieved outstanding results where the data used for training is available in one-to-one pairs. Apart from adversarial loss, it uses L1 (reconstruction) loss as well, which has now become a common practice in these types of tasks.

Unsupervised methods take samples of images from two distributions and learn to cross-translate between them. This introduces the well known issue of there being infinitely many mappings between the two unpaired image domains \cite{cyclegan,cogan,unit,xgan,stargan} and so further constraints are required to do well on the problem. \cite{cyclegan} introduces the requirement that translations be cycle-consistent; mapping image $x \in X$ to domain $Y$ and back again to $X$ must yield an image that is close to the original. \cite{cogan} takes a different approach, enforcing weight sharing between the early layers of the generators and later layers of the discriminators. \cite{unit} combines these two ideas and models each image domain using a VAE-GAN. \cite{xgan} utilizes reconstruction loss and teacher loss instead of VAE using a pretrained teacher network to ensure the encoder output lies in a meaningful subregion. 

To our knowledge, only \cite{stargan} has presented results in generating translations between multiple distribution samples. However, their generator is conditioned on supervised labels.

\section{Method}
\label{ref-method}
Our work broadly builds on the assumption of shared-latent space \cite{cogan}, which theorizes that we can learn a latent code $z$ that can represent the joint distribution $P(x_1, x_2)$, given samples from marginal  distributions $P(x_1)$ and $P(x_2)$. The generator or translator consists of an Encoder $E_i$, shared latent space $z$ and a Decoder $G_i$, such that $z = E_1(x_1)$, $z = E_2(x_2)$ and $x_1 = G_1(z)$, $x_2 = G_2(z)$. The composability property of these translators would be then as follows: $x_2 = G_2 \circ E_1(x_1)$ and vice versa. In other words, we want the translator to learn to map similar characteristics of image samples from two distributions to $z$ and then the decoder should learn to disentangle unique characteristics of that distribution on which it is trained and apply that transformation on any given image sample as input.

We extend this framework to learn composite functions for $|N|$ distributions. To formalize, given sets of samples from distributions $N = \{X_1, X_2, ..., X_{|N|}\}$ with an existing and unknown joint distribution $P(X_1, X_2, ..., X_{|N|}) \neq \phi $, we learn a set of composite functions and a shared latent space, such that $x_j = G_j \circ E_i (x_i)$, where $\{i, j\} \in N$. Thus giving us a total of $|N|^2$ unique transformations possible. To approach solving towards this problem, we start with a bottom-up approach and take $|N| = 4$ sets of sample images.

\subsection{Model architecture}

\begin{figure}[h]
\includegraphics[width=12cm, height=6cm]{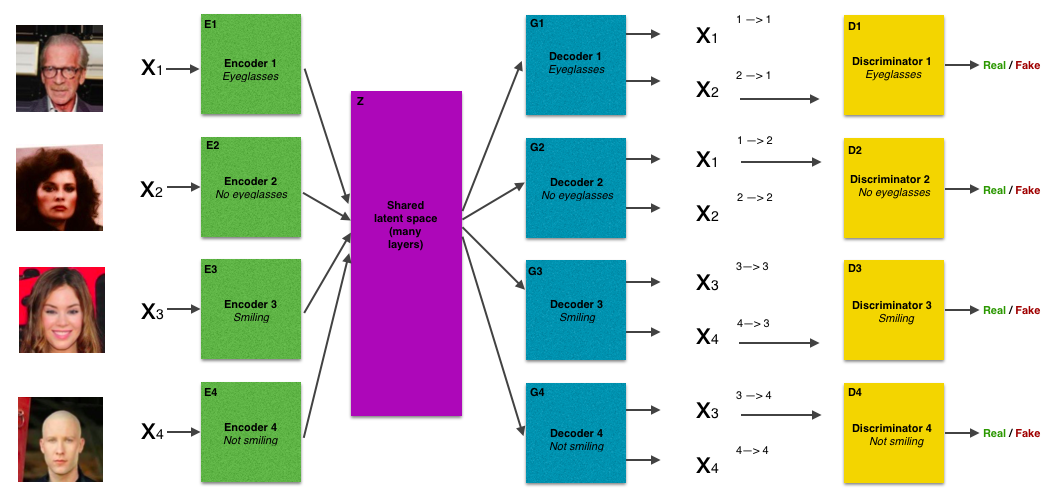}
\centering
\caption{Model architecture. Distributions have been selected from CelebA dataset \cite{celeba} having the following unique properties: smiling, not-smiling, eyeglasses, no-eyeglasses.}
\label{fig-model}
\end{figure}

We extend the model proposed by Liu et. al \cite{unit} to learn to simultaneously translate between two pairs of image distributions (making four distributions in total). There are four encoders, four decoders, and four discriminators in our model, one for each image distribution. Additionally, there is a shared latent space following \cite{unit}, consisting of the last layers of the encoders, and first layers of the decoders. See Figure \ref{fig-model} for more detail. \par

We felt that sharing a latent space, introduced in \cite{cogan}, and used to great effect in \cite{unit} has applications beyond improving pairwise domain translations, and could improve the composability of image translations. Sharing a latent space implements the assumption that there exists a single latent code $z$ from which images in any of the four domains can be recovered \cite{unit}. If this assumption holds, then complex image translations can be disentangled into simpler image translations which learn to map to and from this shared latent code. \par

\subsection{Objective}

We adapted the objective function from \cite{unit}, and benefit from the extensive tuning that the authors carried out. Since we had access to limited computational resources, we kept the same weightings as \cite{unit} on the individual components of loss function for a single pairing. 

There are three components to the objective function for each learned translation, making twelve elements in total.

\begin{align}
&\min_{E_1,E_2,E_3,E_4,G_1,G_2,G_3,G_4}\max_{D_1,D_2,D_3,D_4} \mathcal{L} = \\ &\mathcal{L}_{\text{\tiny VAE}_1}(E_1,G_1) +\mathcal{L}_{\text{\tiny GAN}_1}(E_1,G_1,D_1) +\mathcal{L}_{\text{\tiny CC}_1}(E_1,G_1,E_2,G_2)\nonumber\\
+ &\mathcal{L}_{\text{\tiny VAE}_2}(E_2,G_2) + \mathcal{L}_{\text{\tiny GAN}_2}(E_2,G_2,D_2)+\mathcal{L}_{\text{\tiny CC}_2}(E_2,G_2,E_1,G_1)\nonumber\\
+ &\mathcal{L}_{\text{\tiny VAE}_3}(E_3,G_3) +\mathcal{L}_{\text{\tiny GAN}_3}(E_3,G_3,D_3) +\mathcal{L}_{\text{\tiny CC}_3}(E_3,G_3,E_4,G_4)\nonumber\\
+ &\mathcal{L}_{\text{\tiny VAE}_4}(E_4,G_4) + \mathcal{L}_{\text{\tiny GAN}_4}(E_4,G_4,D_4)+\mathcal{L}_{\text{\tiny CC}_4}(E_4,G_4,E_3,G_3)
\end{align}

The VAE loss objective is responsible for ensuring that the model can reconstruct and image from the same domain. That is,
$$G(E(x)) \approx x$$

The adversarial loss objective is responsible for ensuring that the decoder (or generator $G_i$) generates realistic samples when translating from an image lying in domain $X_1$ into domain $X_2$, which is evaluated by the discriminator ($D_i$). Finally, the cycle-consistency component ensures that when the model translates an image from domain $X_1$ to $X_2$ and back to $X_1$ the resulting image is similar to the original. That is,
$$G_1(E_2(G_2(E_1(x))) \approx x $$

We refer readers to \cite{unit} for a full explanation and motivation of these different elements.

\subsection{Training and Inference}

The model is conceptually split into two, with each part responsible for learning to translate between one pair of distributions. Each of the three loss components; reconstruction, GAN, and cycle-consistency is enforced within the pair.
\begin{enumerate}
    \item (E1,E2,G1,G2,D1,D2): Learns $f_1:X_1 \Rightarrow X_2$, and $f_2:X_2 \Rightarrow X_1$
    \begin{itemize}
        \item $f_1(x) = G_2(E_1(x))$
        \item $f_2(x) = G_1(E_2(x))$
    \end{itemize}
    \item (E3,E4,G3,G4,D3,D4): Learns $f_3:X_3 \Rightarrow X_4$, and $f_4:X_4 \Rightarrow X_3$
    \begin{itemize}
        \item $f_3(x) = G_4(E_3(x))$
        \item $f_4(x) = G_3(E_4(x))$
    \end{itemize}
\end{enumerate}

The shared latent space between all of the encoders and generators is responsible for ensuring realistic translations to image distributions the model has not seen before. \par
At inference time we complete a "double-loop" through the model. Suppose we had learned the following translations:
\begin{itemize}
    \item $f_1$: glasses to no glasses
    \item $f_2$: no glasses to glasses
    \item $f_3$: smiling to not smiling
    \item $f_4$: not smiling to smiling
\end{itemize}
Then to translate from someone who is not smiling and not wearing glasses to smiling and wearing glasses, we do:
\begin{equation}
    \begin{aligned}
        & \text{not smiling, no glasses} \Rightarrow \text{smiling, no glasses} \Rightarrow \text{smiling, glasses} \\
        & \Leftrightarrow f_2(f_4(x)) \\
        & \Leftrightarrow G_2(E_1(G_3(E_4(x))))
    \end{aligned}
\end{equation}

Contrary to the above approach, it seems straightforward to think that training this model would be done in a joint manner with the objective of minimizing $\mathcal{L}$. However, as $N \rightarrow \infty$, this method will become unscalable. Hence we present the above given training strategy that splits the shared latent space $z$ and trains $\frac{|N|}{2}$ pairs.  It must be noted that at this point in the training, there has been no weight sharing between pair 1 ($X_1$ and $X_2$) and pair 2 ($X_3$ and $X_4$). 

In Section \ref{ref:expriments} we see that this method results in generating better quality samples at inference time as compared to joint training from scratch. However, we hypothesized that having a shared latent space would improve the translation quality so experimented with training the models in an uncoupled manner first and then jointly training all the models, sharing a latent space, for a few iterations to fine tune. We found that this approach yielded the best results (see Section \ref{ref:expriments}).

\section{Experiments}
\label{ref:expriments}
We conducted all of our experiments using the celebA dataset \cite{celeba}. This dataset consists of 202,599 images each labeled with 40 binary attributes, for example brown hair, smiling, eyeglasses, beard, and mustache \cite{celeba}. These binary attributes naturally lend themselves to composition, making this an ideal dataset to test our proposed model. We focused on translating between glasses, no glasses, smiling, not smiling (experiment $1$), and blonde hair, brown hair, smiling and not smiling (experiment $2$). For each experiment we constructed four datasets, one corresponding to each image distribution with the relevant characteristic. So that we could test our models for their ability to generate combinations of characteristics that did not appear in the training set, we ensured that there were no faces which were smiling and wearing glasses in experiment $1$, and no faces which were smiling with either blonde or brown hair in experiment $2$. \par

We experimented with the following training approaches.
\begin{itemize}
    \item \textbf{Four way}: Training the model described in Section \ref{ref-method} from scratch
    \item \textbf{Separately Trained (Baseline)}: Following the method and models architectures from \cite{unit}. To compose the image translation we first passed an image through one model, then another.
    \item \textbf{Warm start}: First training separate models. Then initializing the model described in Section \ref{ref-method} with the weights from the separately trained models, and continuing to train to fine tune.
\end{itemize}
The baseline model was intended to help test the role of the shared latent space between all four distributions. If the latent space is helpful, the translations of the four way or warm start model should be better than the baseline model.

High quality translations should have the following characteristics. Realism, variety, and the clear presence of the translated feature, distinct from the pre-translated image. To evaluate our models on these criteria we used three evaluation metrics.
\begin{itemize}
    \item \textbf{Realism}: qualitative, manual examination of the generated images
    \item \textbf{Variety}: Low cycle consistency loss. The lower this loss, the less likely a model is to have mode collapse. If a model experiences mode collapse and translates all example to only a few images, then it will be unable to reconstruct the original image from the translated image well.
    \item \textbf{Presence of translated feature}: We trained a 11-layer VGG \cite{vgg} net using original images from the dataset to classify examples into four classes, one for each possible combination of features for each experiment. Then we selected a batch of 100 original images from a single class (e.g. blonde and not smiling, eyeglasses and not smiling), translated them to every other class using our model, and classified them after every translation. If a model is making clear translations, the class they are classified into should change with each translation. To mitigate the fact that our classifier was imperfect, we excluded any images that the classifier was not able to classify correctly. 
\end{itemize}

\section{Results}

\begin{figure}[h]
\includegraphics[width=14cm]{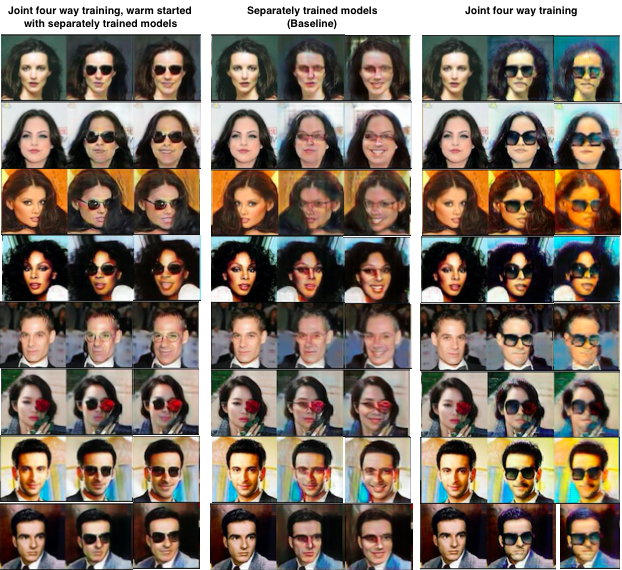}
\centering
\caption{Selected results from experiment 1 for all three models. For each triplet of images, the image on the left is the original image, selected from the celebA dataset \cite{celeba}. They are all not smiling and not wearing glasses. The center image is the translation to not smiling and wearing glasses. The image on the right is the second translation to smiling and wearing glasses.}
\label{fig-eye-smile}
\end{figure}

\begin{figure}[h]
\includegraphics[width=10cm]{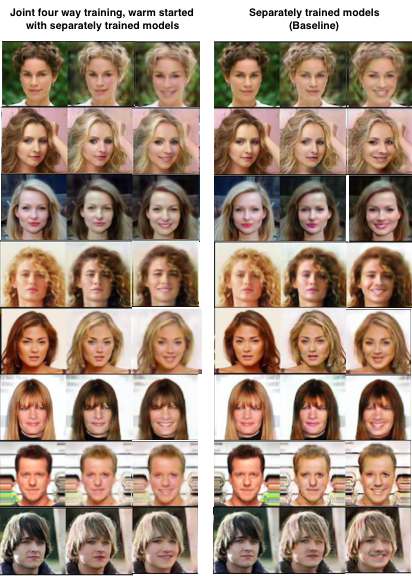}
\centering
\caption{Selected results from experiment 2 for the warm start and baseline models. For each triplet of images, the image on the left is the original image, selected from the celebA dataset \cite{celeba}. They are all not smiling and have either blonde or brown hair. The center image is the translation to not smiling and either blonde or brunette, depending on the original hair color. The image on the right is the second translation to smiling.}
\label{fig-hair-smile}
\end{figure}

\textbf{Realism}: Overall the warm start model described in Section \ref{ref:expriments} generated the most visually appealing and coherent double translations (see Figures \ref{fig-eye-smile} and \ref{fig-hair-smile}). The presence of the translated features are clear and generally integrated in a coherent way, with minimal distortions or artifacts. The model is able to successfully handle atypical translations, such as adding glasses when one eye is occluded (see Figure \ref{fig-eye-smile}). The warm start model is significantly better than a joint model trained from scratch. This is clear from Figure \ref{fig-eye-smile}, and we were not able to successfully train a joint model from scratch for experiment 2, which exhibited results at par with the other methods. Interestingly the separately trained models generated reasonably good double translations, particularly in experiment two (Figure \ref{fig-hair-smile}), and was significantly better than a joint model trained from scratch. This suggests that unpaired image to image translation already exhibits some composablility. However, enforcing a shared latent space and fine tuning these models (the warm start training approach) does seem to improve the overall quality of images. This is particularly apparent in the results from experiment 1 (Figure \ref{fig-eye-smile}). These results suggest that the more scalable decoupled training strategy in which $\frac{|N|}{2}$ pairs are trained separately, then fine-tuned through joint training,  is also the approach which yields the highest quality results. Though we weren't able to experiment on more than 4 distributions due to time and resource constraints. \par
Finally, what is particularly exciting about these results is that our best model has no problem generating images with combinations of characteristics that never appeared in the training set. In experiment 1, there are no pictures of people wearing glasses and smiling, and yet the model generates high quality images of people smiling and wearing glasses (see right most image in the triplets in Figure \ref{fig-eye-smile}). Similarly for experiment 2 there was no one with either brown or blonde hair that was smiling in the training data.

\textbf{Variety}: Generally, our models were able the reconstruct the original image from the translated image well, suggesting they did not suffer from mode collapse. This is also consistent with what we observed by inspecting the generated images.\par

\begin{figure}[h]
\includegraphics[width=14cm]{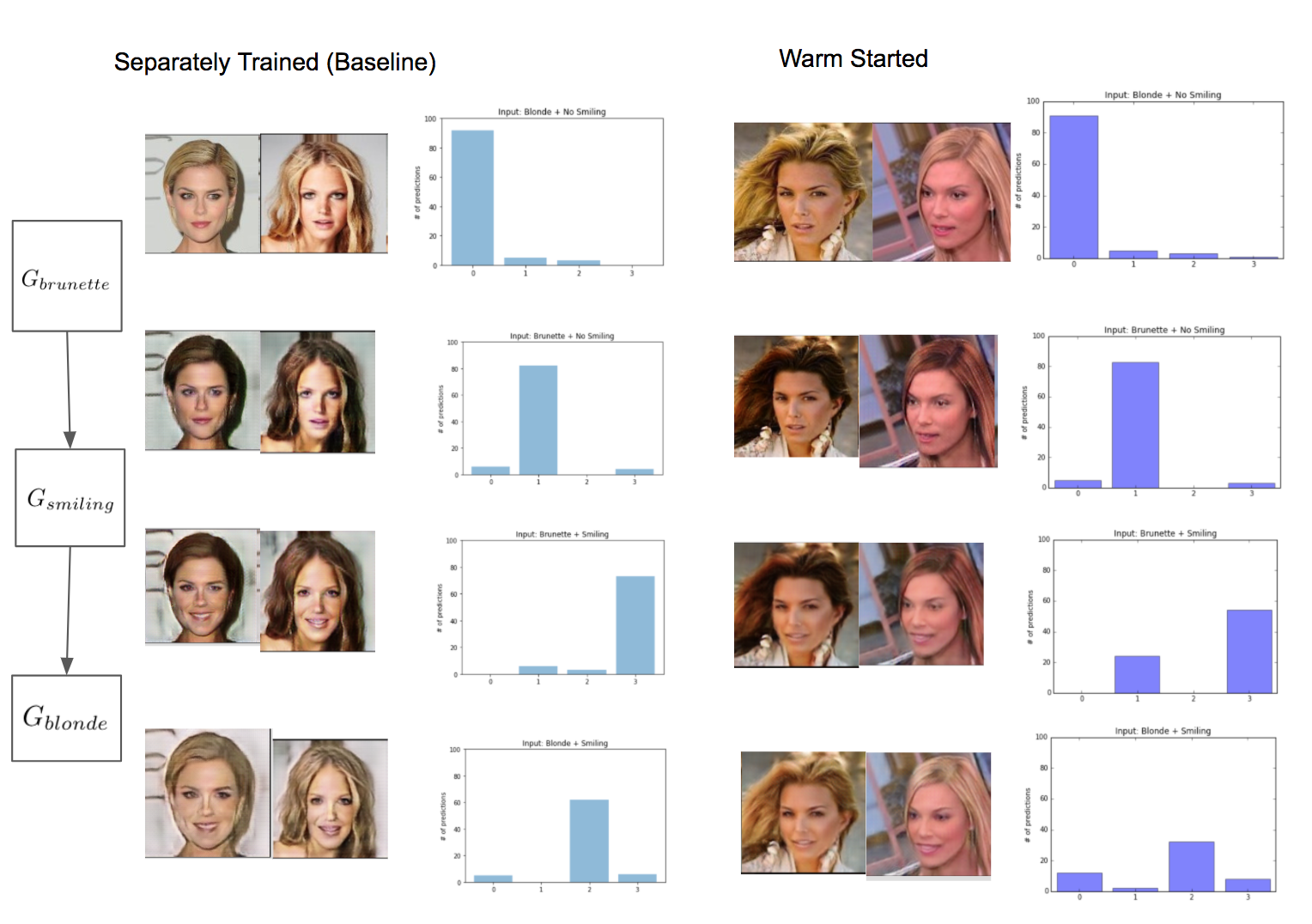}
\centering
\caption{A batch of 100 blonde \& not smiling images are classified and then translated of which again the correctly classified ones are translated again and so on. Some samples from the batch are displayed here. The label map is as follows- 0: Blonde \& Not Smiling, 1: Brunette \& Not Smiling, 2: Blonde \& Smiling, 3: Brunette \& Smiling.}
\label{ref:clf1}
\end{figure}

\textbf{Presence of translated features} (Quantitative Analysis): Figure \ref{ref:clf1} shows our assessment of translation quality. VGG classifier trained on the classes: blonde \& not smiling, brunette \& not smiling, blonde \& smiling, brunette \& smiling  with 87\% accuracy is able to separate the translated images into their respective classes very efficiently for the baseline model. For the warmstarted model, the classifier gets somewhat confused between smiling and not smiling images. This makes us think how finetuning the model is distorting generated samples as to give a mixed classification decision. We leave this for future work.

\section{Further Work}
The joint models we trained sometimes dropped one of the translation modes, most noticeably translating from not smiling to smiling. We hypothesize that this was because this translation was the most difficult of the four translations. This could potentially be remedied by increasing the contribution to the loss function from this translation. More generally it would be interesting to explore the effect of varying the contribution from the many different loss components more fully. Time and computational resource constraints prevented us from doing this.\par
We constructed four images domains from a single more general domain, celebrity faces \cite{celeba}. This ensured that the domains were fundamentally related. It would be interesting to explore the degree of relatedness between different image domains required to achieve good results. For example, given outdoor scenes labeled with the weather (e.g. snow, sun, rain), and outdoor scenes containing different animals (e.g. horse zebra), could we learn to translate between horses in sunshine to zebras in snow? 

\section{Conclusion}
In this work we extend a given model of unpaired image to image translation for handling multiple pairs of distributions. We devise scalable training methods with modified architecture and objective for this kind of model and compare the results of the model through each of these methods. We set qualitative and quantitative evaluation criterion and assess how performance of our model in various training scenarios. Moreover, we show the translation flexibility property of our model by using the translators as stacked composable functions for multi-way translation into novel distributions.

\section*{Acknowledgments}
We are grateful to M. Liu, T. Breuel, and J. Kautz for making their research and codebase publicly available, and to Professor Rob Fergus for his valuable advice.

\bibliography{im2im2im}
\bibliographystyle{ieeetr}

\end{document}